\documentclass[11pt,a4paper]{article}
\usepackage[hyperref]{emnlp2020}
\usepackage{times}
\usepackage{latexsym}

\usepackage{microtype}

\aclfinalcopy 
\def\aclpaperid{2756} 

\usepackage{amsmath, amsfonts, amssymb}
\usepackage{bm}
\usepackage{bbm}
\usepackage{graphicx}
\usepackage{booktabs}
\usepackage{multirow}

\usepackage{array}
\newcolumntype{L}{>{$}l<{$}}
\newcolumntype{C}{>{$}c<{$}}
\newcolumntype{R}{>{$}r<{$}}

\newcommand{\bt}[1]{\textbf{#1}}

\DeclareMathOperator{\bn}{\text{BN}}
\DeclareMathOperator{\linear}{\text{Linear}}
\DeclareMathOperator{\leaky}{\text{LeakyReLU}}
\DeclareMathOperator{\softmax}{\text{Softmax}}

\title{Neural Topic Modeling with Cycle-Consistent Adversarial Training}

\author{Xuemeng Hu\thanks{\; Equal contribution.} \quad Rui Wang\footnotemark[1] \quad 
  Deyu Zhou\thanks{\; Corresponding author.} \quad Yuxuan Xiong \\
  School of Computer Science and Engineering, Key Laboratory of Computer Network \\
  and Information Integration, Ministry of Education, Southeast University, China \\
  {\tt \{xuemenghu,rui\_wang,d.zhou,yuxuanxiong\}@seu.edu.cn} \\}

\date{}

\begin{document}
\maketitle
\begin{abstract}
  Advances on deep generative models have attracted significant research interest in neural topic modeling.
  The recently proposed Adversarial-neural Topic Model models topics with
  an adversarially trained generator network
  and employs Dirichlet prior to capture the semantic patterns in latent topics.
  It is effective in discovering coherent topics but unable to infer topic distributions for given documents
  or utilize available document labels.
  To overcome such limitations, we propose Topic Modeling with Cycle-consistent Adversarial Training (ToMCAT)
  and its supervised version sToMCAT.
  ToMCAT employs a generator network to interpret topics and an encoder network to infer document topics.
  Adversarial training and cycle-consistent constraints are used to
  encourage the generator and the encoder to produce realistic samples that coordinate with each other.
  sToMCAT extends ToMCAT by incorporating document labels
  into the topic modeling process to help discover more coherent topics.
  The effectiveness of the proposed models is evaluated on unsupervised/supervised topic modeling and
  text classification.
  The experimental results show that our models can produce both coherent and informative topics,
  outperforming a number of competitive baselines.
\end{abstract}

\section{Introduction}
Topic models, such as Latent Dirichlet Allocation (LDA) \citep{Blei2003LDA},
aim to discover underlying topics and semantic structures from text collections.
Due to its interpretability and effectiveness, LDA has been extended to many Natural Language Processing (NLP) tasks
\citep{lin2009jst, mcauley2013hidden, zhou-etal-2017-event}.
Most of these models employ mean-field variational inference or
collapsed Gibbs sampling \citep{griffiths2004finding}
for model inference as a result of their intractable posteriors.
However, such inference algorithms are model specific and require dedicated derivations.

To address such limitation,
neural topic models with black-box inference have been explored,
with more flexible training schemes.
Inspired by variational autoencoder (VAE) \citep{kingma2013vae},
\citet{miao2016nvdm} proposed Neural Variational Document Model
which interprets the latent code in VAE as topics.
Following this way,
\citet{srivastava2017prodlda} adopted the logistic normal prior rather than Gaussian to
mimic the simplex properties of topic distribution.
Logistic normal is a Laplace approximation to the Dirichlet distribution \citep{MacKay1998laplace}.
However, logistic normal can not exhibit multiple peaks at the vertices of the simplex as the Dirichlet distribution.
Therefore, it is less capable of capturing
the multi-modality which is crucial for topic modeling \citep{Wallach2009ldaprior}.

To overcome such limitation,
\citet{wang2019atm} proposed Adversarial-neural Topic Model (ATM),
a topic model based on Generative Adversarial Networks (GANs) \citep{goodfellow2014gan}
and sampling topics directly from the Dirichlet distribution to impose a Dirichlet prior.
ATM employs a generator transforming randomly sampled topic distributions to word distributions,
and an adversarially trained discriminator estimating the probability
that a word distribution came from the training data rather than the generator.
Although ATM was shown to be effective in discovering coherent topics,
it can not be used to induce the topic distribution given a document
due to the absence of a topic inference module.
Such limitation hinders its application to downstream tasks, such as text classification.
Moreover, ATM fails to deal with document labels which can help extract more coherent topics.
For example, a document labeled as \emph{`sports'}
more likely belongs to topics such as \emph{`basketball'} or \emph{`football'}
rather than \emph{`economics'} or \emph{`politics'}.

To address such limitations of ATM,
we propose a novel neural topic modeling approach,
named Topic Modeling with Cycle-consistent Adversarial Training (ToMCAT).
In ToMCAT, topic modeling is cast into the transformation between topic distributions and word distributions.
Specifically, the transformation from topic distributions to word distributions is used to interpret topics,
and the reverse transformation
is used to infer underlying topics for a given document.
Under such formulation, ToMCAT employs
a generator to transform topic distributions randomly sampled from the Dirichlet prior
into the corresponding word distributions,
and an encoder to reversely transform documents represented as word distributions into their topic distributions.
To encourage the generator/encoder to produce more realistic target samples,
discriminators for word/topic distributions are introduced to enable adversarial training.
Additional cycle-consistency constraints are utilized to align
the learning of the encoder and the generator to prevent them from contradicting each other.
Furthermore, for documents with labels,
we propose sToMCAT that introduces an extra classifier to regularize the topic modeling process.

The main contributions of the paper are:
\begin{itemize}
  \item ToMCAT, a novel topic model with cycle-consistent adversarial training is proposed.
        To the best of our knowledge, it is the first adversarial topic modeling approach
        with both topic discovery and topic inference.
  \item sToMCAT, a supervised extension to ToMCAT, is proposed
        to help discover more coherent topics with available document labels.
  \item Experimental results on unsupervised/super-vised topic modeling and text classification
        demonstrate the effectiveness of the proposed approaches.
\end{itemize}

\section{Related Work}\label{sec:related}
Our work is related to neural topic modeling and unsupervised style transfer.

\subsection{Neural Topic Modeling}
Recent advances on deep generative models,
such as VAEs \cite{kingma2013vae} and GANs \cite{goodfellow2014gan},
attract much research interest in the NLP community.

Based on VAE, Neural Variational Document Model (NVDM) \citep{miao2016nvdm}
encodes documents with variational posteriors in the latent topic space.
NVDM employs Gaussian as the prior distribution of latent topics.
Instead,
\citet{srivastava2017prodlda} proposed that Dirichlet distribution is
a more appropriate prior for multinomial topic distributions,
and constructed a Laplace approximation of Dirichlet
to enable reparameterisation \citep{kingma2013vae}.
Furthermore, the word-level mixture is replaced with a weighted product of experts \citep{srivastava2017prodlda}.
Later, a non-parametric neural topic model utilizing stick-breaking construction
was presented in \citep{miao2017discovering}. There are some attempts in
incorporating supervised information into neural topic modeling.
For example, \citet{card-etal-2018-neural} extended
the Sparse Additive Generative Model \citep{eisenstein2011sparse} in the neural framework
and incorporated document metadata such as document labels into the modeling process.

Apart from VAE-based approaches, Adversarial-neural Topic Model (ATM) \citep{wang2019atm})
was proposed to model topics with GANs.
The generator of ATM projects randomly sampled topic distributions
to word distributions, and is adversarially trained with a discriminator that tries to
distinguish real and generated word distributions.
Moreover, \citet{wang-etal-2019-open} extended ATM for open-domain event extraction by
representing an event as a combination of an entity distribution, a location distribution,
a keyword distribution and a date distribution.
Such joint distributions are adversarially learned in a similar manner as ATM.
The proposed ToMCAT is partly inspired by ATM but differs in its capability of
inferring document-specific topic distributions and incorporating supervision information.
BAT \citep{wang-etal-2020-neural} is an extension to ATM that employs
bidirectional adversarial training \citep{donahue2016adversarial}
for document-specific topic distribution inference.
Although BAT similarly utilizes an adversarial training objective to guide the learning of topic distribution,
there are some major differences.
Apart from different adversarial losses, ToMCAT also incorporates two cycle-consistency constraints
which encourage the model to generate informative representations
and are shown to be crucial for generating coherent topics as in our experiments.

\subsection{Unsupervised Style Transfer}
Style transfer, aiming at transforming representations from one style to another,
has been found many interesting applications, such as image and text style transfer.
However, training data paired between different styles are not available for many tasks.
To solve this problem, \citet{zhu2017cycle} imposed cycle-consistency constraints
to align mappings between two styles and proposed CycleGAN for unsupervised image style translation.
Similarly, DiscoGAN \citep{kim2017learning} was proposed to discover the relations between different image styles
and transformed images from one style to another without paired data.
In the NLP field, \citet{lee2018scalable} developed a CycleGAN-based approach
to transfer the sentiment style (positive, negative) of the text.

Inspired by CycleGAN,
Our work views topic modeling as unsupervised distribution transfer and follows the framework of CycleGAN.

\section{Methodology}\label{sec:methodology}

Given a corpus $\mathcal{D}$ consisting of $N$ documents $\{\bm{x}_i\}_{i=1}^{N}$,
two main purposes of topic modeling are:
\begin{enumerate}
  \item Topic {\color{black}discovery}.
        Given a one-hot topic indicating vector $\bm{I}_k \in \mathbb{R}^K$
        where $K$ is the number of topics and $I_{kk}=1$,
        discover the corresponding word distribution $\bm{t}_k \in \mathbb{R}^V$ from $\mathcal{D}$
        where $V$ is the vocabulary size.
        More generally, we can consider topic discovery as finding a mapping from topic distribution
        to word distribution.
  \item Topic inference. Infer the topic distribution $\bm{z}_j \in \mathbb{R}^K$
        of the document $\bm{x}_j \in \mathbb{R}^V$.
        Similarly, the topic inference can be considered as finding a mapping
        from word distribution to topic distribution.
\end{enumerate}

We now formalize the above observations.
Let $X$ be the word distribution set and $Z$ the topic distribution set.
Given training samples $\{\bm{x}_i\}_{i=1}^{N}$ where $\bm{x}_i \in X$ and
document-specific topic distributions $\{\bm{z}_j\}_{j=1}^M$ where $\bm{z}_j \in Z$,
the goal of topic modeling is to learn a mapping function $G$, called \emph{generator},
to transform samples in $Z$ into $X$ and a reverse function $E$, called \emph{encoder},
to transform samples in $X$ into $Z$.
However, it should be noted that training samples in $X$ and $Z$ are unpaired
since the topic distribution of a document is unknown before topic modeling.
Thus, the problem is how to learn $G$ and $E$ to model topics
in the absence of paired samples between $X$ and $Z$.

\subsection{ToMCAT}

\begin{figure}[t]
  \centering
  \resizebox{\columnwidth}{!}{
    \includegraphics[]{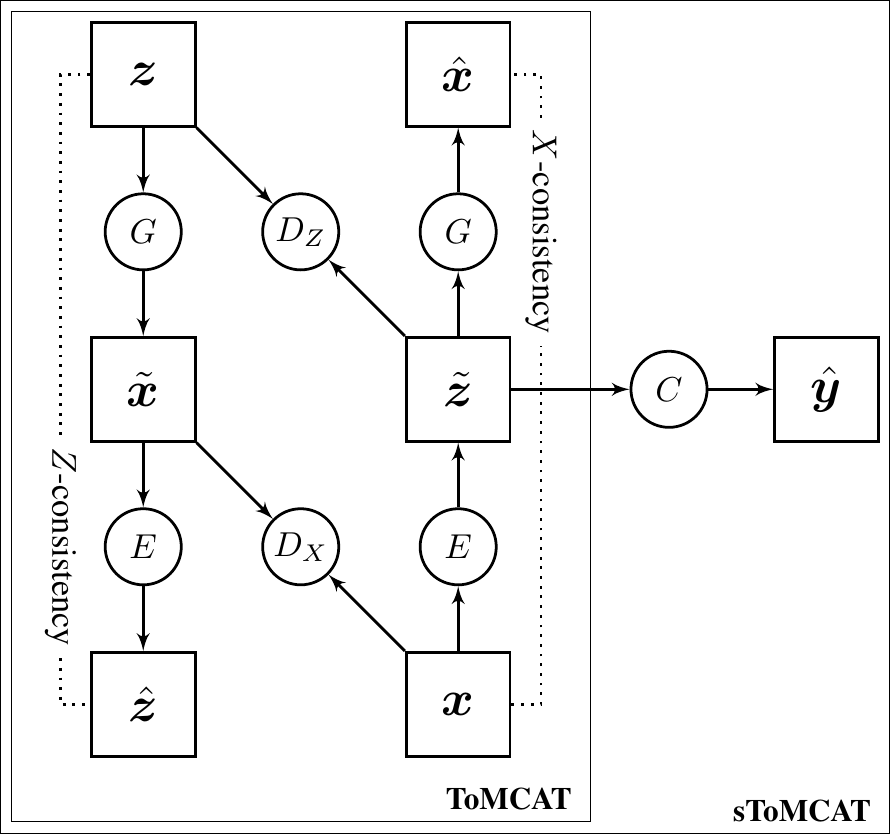}
  }
  \caption{The framework of ToMCAT and sToMCAT.
    Circles are neural networks,
    squares are data representations,
    and arrows indicate the forward pass directions.
  }
  \label{fig:framework}
\end{figure}

We now introduce the proposed ToMCAT, which is
shown in the inner panel of \mbox{Figure \ref{fig:framework}}.

ToMCAT consists of a generator $G$: $Z \rightarrow X$, an encoder $E$: $X \rightarrow Z$,
and adversarial discriminators $D_X$ and $D_Z$ of $G$ and $E$ respectively. Following CycleGAN \citep{zhu2017cycle},
ToMCAT employs two types of losses, namely adversarial losses and cycle-consistency losses,
to guide the training of the encoder $E$ and the generator $G$.
The details of these modules are described below.

\subsubsection{Encoder Network $E$}
Encoder $E$ transforms a word distribution $\bm{x}_i \in \mathbb{R}^V$ into
its corresponding topic distribution $\bm{z}_i \in \mathbb{R}^K$.
Following \cite{wang2019atm},
we represent $\bm{x}_i \in X$ with the normalized TF-IDF (Term Frequency–Inverse Document Frequency) representation of $i$-th document:
\begin{align}
  \hat d_{ij} & = \frac{d_{ij}}{\sum_j d_{ij}} \cdot \log
  \frac{N}{1 + \sum_{n=1}^N \mathbbm{1}(d_{nj} \neq 0)} \label{eq:tfidf},        \\
  x_{ij}      & = \frac{\hat d_{ij}}{\sum_{v} \hat d_{iv}} \label{eq:tfidfNorm},
\end{align}
where $d_{ij}$ is the count of $j$-th word in $i$-th document,
$\mathbbm{1}(\cdot)$ denotes the indicator function.
Equation \ref{eq:tfidf} calculates the smoothed TF-IDF of $\bm{d}_i$,
which is then normalized to sum to one in Equation \ref{eq:tfidfNorm}.
We use TF-IDF as the document representation because
TF-IDF generally preserves the relative importance of words in a document
and reduce the noise of stop words.
As the target distribution of the generator $G$, such property of TF-IDF will help generate more informative topics.

The implementation of the encoder is
a multilayer perception (MLP) with LeakyReLU activation
\citep{maas2013leaky} and batch normalization (BN) \citep{Ioffe2015bn}.
The detailed transformations are:
$[\,\linear(V, H)$ $\rightarrow$ $\leaky(0.1)$ $\rightarrow$ $\bn$
      $\rightarrow$ $\linear(H, K)$ $\rightarrow$ $\softmax \,]$,
where $\linear(I, J)$ denotes a linear transformation from $I$-dim to $J$-dim,
$H$ is the number of hidden units,
and the final $\softmax$ makes sure that the final output is one-normalized
to match the input of the generator $G$.
Inputs of $E$ are either sampled from the corpus $\mathcal{D}$ or generated by $G$.

\subsubsection{Generator Network $G$}
The generator $G$ performs the reverse operation of the encoder by
transforming a topic distribution $\bm{z}_j \in \mathbb{R}^K$
into a word distribution $\bm{x}_j \in \mathbb{R}^V$,
where the input $\bm{z}_j$ is generated by the encoder or sampled from the prior distribution.
To draw the topic distribution $\bm{z}_j$,
a common practice for topic modeling
is to use the Dirichlet distribution, the conjugate prior of the multinomial distribution.
We also stick with this choice in our model.
Specifically,
we draw topic distributions from a symmetric Dirichlet distribution with parameters
$\bm{A} \in \mathbb{R}^K$ where $\bm{A}_k = \alpha$ for $1 \leq k \leq K$.

After sampling a topic distribution $\bm{z}_j$ from the Dirichlet prior,
the generator then maps $\bm{z}_j$ from $Z$ to $X$,
and the transformations is similar to the encoder:
$[\,\linear(K, H)$ $\rightarrow$ $\leaky(0.1)$ $\rightarrow$ $\bn$
      $\rightarrow$ $\linear(H, V)$ $\rightarrow$ $\softmax \,]$,
where the final output is also normalized by the $\softmax$
to match the input of the encoder.

\subsubsection{Training Objective}
Following CycleGAN \citep{zhu2017cycle},
we employ adversarial losses and cycle-consistency losses to guide the training of $G$ and $E$.
The adversarial losses encourage $G$ and $E$ to generate samples matching
the data distribution in the target space ($X$ for $G$ and $Z$ for $E$)
while the cycle-consistency losses align $G$ and $E$ in these two distribution spaces
to prevent them from contradicting each other.

\paragraph{Adversarial Loss}
Generator $G$ is adversarially trained with a discriminator $D_X$,
which takes as input either real samples from training data,
i.e., $\bm{x} \sim p_{\text{data}}(\bm{x})$, or fake samples generated by $G$, i.e., $G(\bm{z})$.
The goal of $D_X$ is to distinguish real samples from fake ones,
while $G$ instead aims to fool $D_X$ by generating samples similar to $\bm{x}$.
Therefore, the adversarial training encourages $G$ to mimic the pattern of $X$
and produce realistic word distributions.
We employ a Wasserstein GAN (WGAN) \citep{arjrjovsky2017wgan} based adversarial loss to $G$ and $D_X$:
\begin{align}\label{eq:wganLoss}
  \mathcal{L}_{\text{adv}}(G, D_X) = {}
   & \mathbb{E}_{\bm{x} \sim p_{\text{data}}(\bm{x})}[D_X(\bm{x})] - \notag \\
   & \mathbb{E}_{\bm{z} \sim p_{\text{data}}(\bm{z})}[D_X(G(\bm{z}))],
\end{align}
where $D$ tries to maximize $\mathcal{L}_{\text{adv}}(G, D_X)$ while $G$ tries to minimize it.
Similarly, the adversarial loss applied to $E$ and $D_Z$ is:
\begin{align}\label{eq:wganLossZ}
  \mathcal{L}_{\text{adv}}(E, D_Z) = {}
   & \mathbb{E}_{\bm{z} \sim p_{\text{data}}(\bm{z})}[D_Z(\bm{z})] - \notag \\
   & \mathbb{E}_{\bm{x} \sim p_{\text{data}}(\bm{x})}[D_Z(E(\bm{x}))].
\end{align}

Discriminators $D_X$ and $D_Z$ are implemented with MLPs,
and we use the same architecture for them: $[\,\linear(S, H)$ $\rightarrow$ $\leaky(0.1)$ $\rightarrow$ $\bn$
      $\rightarrow$ $\linear(H, 1) \,]$,
where $S$ equals to $V$ for $D_X$ and $K$ for $D_Z$.
Since we are using WGAN rather than
the original GAN loss as in CycleGAN,
we do not apply a sigmoid transformation to discriminator outputs.

\paragraph{Cycle-Consistency Loss}
Adversarial training might lead to generating samples
identically distributed as corresponding target samples \citep{goodfellow2014gan}.
However, the relationship between the source distributions and the transformed distributions is unconstrained.
\citet{zhu2017cycle} argued that adversarial losses alone is not able to fulfill this task and
that the learned mappings should be cycle-consistent to reduce the search space of possible mapping functions,
i.e., $\bm{x} \rightarrow E(\bm{x}) \rightarrow G(E(\bm{x})) \approx \bm{x}$ and
$\bm{z} \rightarrow G(\bm{z}) \rightarrow E(G(\bm{z})) \approx \bm{z}$.
To this end, two cycle-consistency losses
$\overrightarrow {\mathcal{L}_{\text{cyc}}} (G, E)$
and $\overleftarrow {\mathcal{L}_{\text{cyc}}} (G,E)$
are added to the training objective,
as shown in the inner panel (dotted lines) of Figure \ref{fig:framework}. Specifically,
\begin{equation}\label{eq:cycLoss}
  \begin{aligned}
    \overrightarrow {\mathcal{L}_{\text{cyc}}} (G, E)=
    \mathbb{E}_{\bm{x} \sim p_{\text{data}}(\bm{x})}[\lVert G(E(\bm{x})) - \bm{x} \rVert _1], \\
    \overleftarrow {\mathcal{L}_{\text{cyc}}} (G,E)=
    \mathbb{E}_{\bm{z} \sim p_{\text{data}}(\bm{z})}[\lVert E(G(\bm{z})) - \bm{z} \rVert _ 1],
  \end{aligned}
\end{equation}
where $\lVert \cdot \rVert _1$ denotes L1 norm.

\paragraph{Overall Objective}
Summing up adversarial losses in Equation \ref{eq:wganLoss}, \ref{eq:wganLossZ}
and cycle-consistency losses in Equation \ref{eq:cycLoss},
the overall objective of ToMCAT is:
\begin{multline}\label{eq:cycleFullLoss1}
  \mathcal{L}(G, E, D_X, D_Z) = \\
  \mathcal{L}_{\text{adv}}(G, D_X) + \mathcal{L}_{\text{adv}}(E, D_Z) + \\
  \lambda_1 \overrightarrow {\mathcal{L}_{\text{cyc}}}(G, E) +
  \lambda_2 \overleftarrow {\mathcal{L}_{\text{cyc}}}(G, E),
\end{multline}
where $\lambda_1$ and $\lambda_2$ respectively control the relative importance of
$\overrightarrow {\mathcal{L}_{\text{cyc}}}(G, E)$ and $\overleftarrow {\mathcal{L}_{\text{cyc}}}(G, E)$
w.r.t. adversarial losses.

\subsection{sToMCAT}
The encoder $E$ transforms the word distribution $\bm{x}$ to corresponding topic distribution $\bm{z}$,
which effectively captures the key semantic information of $\bm{x}$ and can be directly used to downstream tasks,
e.g., text classification.
Therefore, for labeled documents we extend ToMCAT with a classifier $C$ to
allow the incorporation of label information, as shown in Figure \ref{fig:framework}.
We name the supervised version as sToMCAT.

For a word distribution $\bm{x}$ and its one-hot label $\bm{y}$,
$\bm{x}$ is first encoded by the encoder $E$ into the topic distribution $\bm{z}$,
and then $\bm{z}$ is fed to the classifier $C$ to predict the probability of $\bm{y}$.
The predictive objective is defined as:
\begingroup\makeatletter\def\f@size{10}\check@mathfonts
\begin{equation}\label{eq:clsLoss}
  {\mathcal{L}_{\text{cls}}} (E, C)= \\
  - \mathbb{E}_{(\bm{x}, \bm{y}) \sim p_{\text{data}}(\bm{x}, \bm{y})}[\bm{y} \log C(E(\bm{x}))],
\end{equation}
\endgroup
where $L$ is the dimension of $\bm{y}$.
We employ an MLP classifier:
$[\,\linear(K, H)$ $\rightarrow$ $\leaky(0.1)$ $\rightarrow$ $\bn$
      $\rightarrow$ $\linear(H, L)$ $\rightarrow$ $\softmax \,]$.

For sToMCAT, the topic model and the classifier are trained jointly, and its overall objective is defined as:
\begin{multline}\label{eq:cycleFullLossSup}
  \mathcal{L}\text{sup}(G, E, D_X, D_Z, C) = \\
  \mathcal{L}(G, E, D_X, D_Z) + \lambda_3 \mathcal{L}_{\text{cls}}(E, C).
\end{multline}

\subsection{Training Details}
The proposed ToMCAT and sToMCAT are trained with the Adam optimizer \citep{kingma2014adam},
whose learning rate and momentum term $\beta_1$ are set to $0.0001$ and $0.5$ respectively
for ($G$, $E$, $D_X$) and $D_Z$,
while $0.001$ and $0.9$ for the classifier $C$.
The hidden unit numbers are set to $100$ for all modules.
Besides, to enforce the Lipschitz constraints required by WGAN,
a weight clipping of $0.01$ is adopted \citep{arjrjovsky2017wgan}.
\footnote{We also experiment with the gradient-penalty WGAN \citep{gulrajani2017wgangp},
  but the weight clipping version performs better in general.}

During training, the parameters of discriminators $D_X$, $D_Z$ and
mappings $G$, $E$ are alternately updated.
Specifically, at each training iteration, firstly we optimize $D_X$ and $D_Z$ for $5$ steps with adversarial losses,
and then another training step is taken to optimize $G$ and $E$ with adversarial losses and cycle-consistency losses
(Equation \ref{eq:cycleFullLoss1}).
When the model is trained in a supervised way,
the predictive objective is additionally applied to $E$ and $C$ at the last training step
(Equation \ref{eq:cycleFullLossSup}).

We found that relatively good choices of $\lambda_1$ and $\lambda_2$ fall into different regions
for different datasets and topic number settings,
which implies a further tuning of these hyperparameters is needed.
To ease this kind of burden, we apply a gradient-based mechanism to adversarial losses and cycle-consistency losses.
It balances these two types of losses
with the L2 norms of their gradients w.r.t. the output of their preceding mapping functions.
E.g., for $\mathcal{L}_{\text{adv}}(G, D_X)$ and $\overrightarrow {\mathcal{L}_{\text{cyc}}}(G, E)$,
we replace $\lambda_1$ in Equation \ref{eq:cycleFullLoss1} with:
\begin{equation}\label{eqLambda}
  \lambda _1 = \hat \lambda_1 \frac
  {\lVert \partial \mathcal{L}_{\text{adv}}(G, D_X) / \partial G(\bm{z}) \rVert _2}
  {\lVert \partial \overrightarrow {\mathcal{L}_{\text{cyc}}}(G, E) / \partial G(\bm{z}) \rVert _2},
\end{equation}
where $\hat \lambda_1$ is the new balancing factor and $\lVert \cdot \rVert _2$ denotes L2 norm.
Similarly, $\mathcal{L}_{\text{adv}}(E, D_Z)$ and $\overleftarrow {\mathcal{L}_{\text{cyc}}}(G, E)$,
$\mathcal{L}_{\text{adv}}(E, D_Z)$ and $\mathcal{L}_{\text{cls}}(E, C)$ are also balanced in this way
with $\hat \lambda_2$ and $\hat \lambda_3$.
The resulting $\hat \lambda_1$, $\hat \lambda_2$ and $\hat \lambda_3$ are set to $2$, $0.2$ and $1$ respectively
for all datasets and topic number settings in our experiments,
thus avoiding the  time-consuming hyperparameter tuning process.

\section{Experiments}\label{sec:exp}
In this section, we first describe datasets and compared baselines.
Then we present topic modeling results under both unsupervised and supervised settings.
Finally, we report the text classification results.

\begin{table}[ht]
  \small
  \centering
  \resizebox{\columnwidth}{!}{
    \begin{tabular}{lrrrr}
      \toprule
      Dataset & \#Train & \#Test & Vocab Size & \#Class \\
      \midrule
      NYT     & 99,992  & -      & 12,604     & -       \\
      GRL     & 29,762  & -      & 15,276     & -       \\
      DBP     & 99,991  & 69,993 & 9,005      & 14      \\
      20NG    & 11,258  & 7,492  & 2,000      & 20      \\
      \bottomrule
    \end{tabular}
  }
  \caption{Dataset statistics.}
  \label{tb:stat}
\end{table}

\begin{table*}[t]
  \centering
  \resizebox{2.1\columnwidth}{!}{
    \begin{tabular}{llRRRRRRRRRR}
      \toprule
      \multirow{2}{*}{Dataset}
       &
      \multirow{2}{*}{Metric}
       &
      \multicolumn{7}{c}{Unsupervised}
       &
      \multicolumn{3}{c}{Supervised}                                                                                 \\
      \cmidrule(lr){3-9} \cmidrule(lr){10-12}
       &
       &
      \multicolumn{1}{c}{NVDM}
       &
      \multicolumn{1}{c}{ProdLDA}
       &
      \multicolumn{1}{c}{ATM}
       &
      \multicolumn{1}{c}{BAT}
       &
      \multicolumn{1}{c}{LDA}
       &
      \multicolumn{1}{c}{Scholar}
       &
      \multicolumn{1}{c}{ToMCAT}
       &
      \multicolumn{1}{c}{sLDA}
       &
      \multicolumn{1}{c}{Scholar}
       &
      \multicolumn{1}{c}{sToMCAT}                                                                                    \\
      \midrule
      \multirow{3}{*}{NYT}
       & C\_A & 0.0770  & 0.1841  & 0.2292 & 0.2356 & 0.2145 & 0.1949  & \bm{0.2444} & -      & -      & -           \\
       & C\_P & -0.5368 & 0.1255  & 0.3330 & 0.3749 & 0.3230 & 0.0451  & \bm{0.3879} & -      & -      & -           \\
       & NPMI & -0.1461 & 0.0155  & 0.0806 & 0.0952 & 0.0814 & -0.0290 & \bm{0.0956} & -      & -      & -           \\
      \midrule
      \multirow{3}{*}{GRL}
       & C\_A & 0.0715  & 0.1483  & 0.2203 & 0.2108 & 0.1960 & 0.2064  & \bm{0.2285} & -      & -      & -           \\
       & C\_P & -0.5188 & -0.0651 & 0.2576 & 0.2312 & 0.1974 & 0.2150  & \bm{0.2752} & -      & -      & -           \\
       & NPMI & -0.1225 & -0.0193 & 0.0655 & 0.0608 & 0.0533 & 0.0592  & \bm{0.0808} & -      & -      & -           \\
      \midrule
      \multirow{3}{*}{DBP}
       & C\_A & 0.1385  & 0.2653  & 0.2928 & 0.2355 & 0.2756 & 0.3010  & \bm{0.3410} & 0.2216 & 0.2966 & \bm{0.3568} \\
       & C\_P & -0.2970 & 0.2149  & 0.3397 & 0.3749 & 0.3516 & 0.2369  & \bm{0.4327} & 0.2581 & 0.1834 & \bm{0.4981} \\
       & NPMI & -0.1171 & 0.0212  & 0.1100 & 0.0951 & 0.1033 & 0.0661  & \bm{0.1434} & 0.0685 & 0.0526 & \bm{0.1661} \\
      \midrule
      \multirow{3}{*}{20NG}
       & C\_A & 0.1115  & 0.1776  & 0.1833 & 0.1991 & 0.1862 & 0.1777  & \bm{0.2082} & 0.1771 & 0.1811 & \bm{0.2248} \\
       & C\_P & -0.0632 & 0.0709  & 0.2572 & 0.2962 & 0.2816 & 0.2120  & \bm{0.3137} & 0.2621 & 0.2443 & \bm{0.3563} \\
       & NPMI & -0.0495 & -0.0439 & 0.0379 & 0.0555 & 0.0637 & 0.0426  & \bm{0.0656} & 0.0554 & 0.0486 & \bm{0.0709} \\
      \bottomrule
    \end{tabular}
  }
  \caption{Average topic coherence of 5 topic number settings (20, 30, 50, 75, 100) on 4 datasets.
    Bold values indicate the best performing models
    for each dataset/metric/supervision setting. }
  \label{tb:coherence}
\end{table*}

\subsection{Experimental Setup}\label{ssec:expSetup}
We evaluate the performance of proposed models on four datasets:
NYTimes\footnote{\url{http://archive.ics.uci.edu/ml/datasets/Bag+of+Words}} (\textbf{NYT}),
Grolier\footnote{\url{https://cs.nyu.edu/~roweis/data}} (\textbf{GRL}),
DBpedia ontology classification dataset (\textbf{DBP}) \citep{zhang2015charconv} and
20 Newsgroups\footnote{\url{http://qwone.com/~jason/20Newsgroups}} (\textbf{20NG}).
For NYTimes and Grolier datasets, we use the processed version of \citep{wang2019atm}.
For the DBpedia dataset, we first sample $100,000$ documents from the whole training set, and then perform
preprocessing including tokenization, lemmatization, removal of stopwords, and low-frequency words.
The same preprocessing is also applied to the 20 Newsgroups dataset.
The statistics of the processed datasets are shown in Table \ref{tb:stat}.

We choose the following approaches as our baselines:
\vspace{-\topsep}
\begin{itemize}
  \itemsep0em
        \setlength{\itemsep}{1pt}
        \setlength{\parskip}{0pt}
        \setlength{\parsep}{0pt}
  \item LDA \citep{Blei2003LDA}.
        We use GibbsLDA++, an implementation using Gibbs sampling for parameter estimation
        and inference\footnote{\url{http://gibbslda.sourceforge.net}}.
  \item sLDA\footnote{\url{https://github.com/blei-lab/class-slda}} \citep{mcauliffe2008slda},
        a supervised extension to LDA.
  \item NVDM\footnote{\url{https://github.com/ysmiao/nvdm}} \citep{miao2016nvdm}, a VAE-based model that
        employs Gaussian prior for the latent topics.
  \item ProdLDA\footnote{\url{https://github.com/akashgit/autoencoding_vi_for_topic_models}}
        \citep{srivastava2017prodlda}, a VAE-based model
        that replaces the mixture model with a product of experts.
  \item Scholar\footnote{\url{https://github.com/dallascard/scholar}}  \citep{card-etal-2018-neural},
        a ProdLDA-based model that enables optional incorporation of metadata.
  \item ATM \citep{wang2019atm}, a neural topic model utilizing adversarial training.
  \item BAT \citep{wang-etal-2020-neural}, a neural topic model utilizing bidirectional adversarial training.
\end{itemize}

\subsection{Topic Modeling}\label{ssec:expTopic}
We evaluate the performance of the proposed models and baselines using topic coherence measures.
Topic coherence measures are metrics for quantifying the understandability of the extracted topics,
which are shown highly correlated with human subjects
\citep{newman-etal-2010-automatic, aletras-stevenson-2013-evaluating}.
Since a topic is typically represented as a word distribution over the vocabulary or
$n$ top-weighted words (i.e., topic words) in this distribution,
we calculate the coherence of a topic by measuring the relatedness between its topic words.
The word relatedness scores are estimated based on some kind of word co-occurrence statistics on Wikipedia,
for example, by applying a sliding window over the Wikipedia corpus
and collecting word co-occurrences to calculate
NPMI (Normalized Pointwise Mutual Information) \citep{bouma2009normalized} for word pairs.
We refer readers to \citep{roder2015exploring} for detailed calculation and comparison
of different topic coherence measures.
In our experiments, we use top-10 topic words of each topic to calculate topic coherence
and report the results of 3 topic coherence measures:
C\_A \citep{aletras-stevenson-2013-evaluating},
C\_P \citep{roder2015exploring},
and NPMI \citep{aletras-stevenson-2013-evaluating}.
The topic coherence scores are calculated using Palmetto
\footnote{\url{https://github.com/AKSW/Palmetto}}.

\subsubsection{Unsupervised Topic Modeling}
To make a more comprehensive comparison of our model with baselines for topic modeling,
we experiment on each dataset with five topic number settings: 20, 30, 50, 75, 100.
The average topic coherence scores of 5 settings are presented in Table \ref{tb:coherence}.
We can see from the left part of Table \ref{tb:coherence} that, among all unsupervised topic models,
our model achieves the highest scores on all datasets and topic coherence measures.

With an improper Gaussian prior,
NVDM shows the worst performance among all neural topic models with no exception.
The logistic-normal based ProdLDA and Scholar achieve higher topic coherence scores compared to NVDM,
but are still largely underperformed compared to our model.
BAT achieves the second-best place most of the time in unsupervised topic modeling experiments.
Compared to ToMCAT, BAT has a similar adversarial objective but lacks the cycle-consistency constraints,
Therefore, the generator and encoder of BAT only aim to fool the discriminator by mimicking the pattern of
the joint distribution of real documents and topics.
With the incorporation of two cycle-consistency losses, ToMCAT is explicitly encouraged to
generate not only realistic but also informative representations in order to
reduce the cycle-consistency losses.

To give an insight into the generated topics,
8 out of 50 topics discovered by ToMCAT on NYTimes are presented in Table \ref{tb:topic8},
where a topic is represented by the ten words with the highest probability in the topic.
We can observe that the extracted topics are highly coherent and interpretable.
The corresponding full list of topics can be found in the appendix.

\begin{figure*}[!ht]
  \centering
  \includegraphics[width=\textwidth]{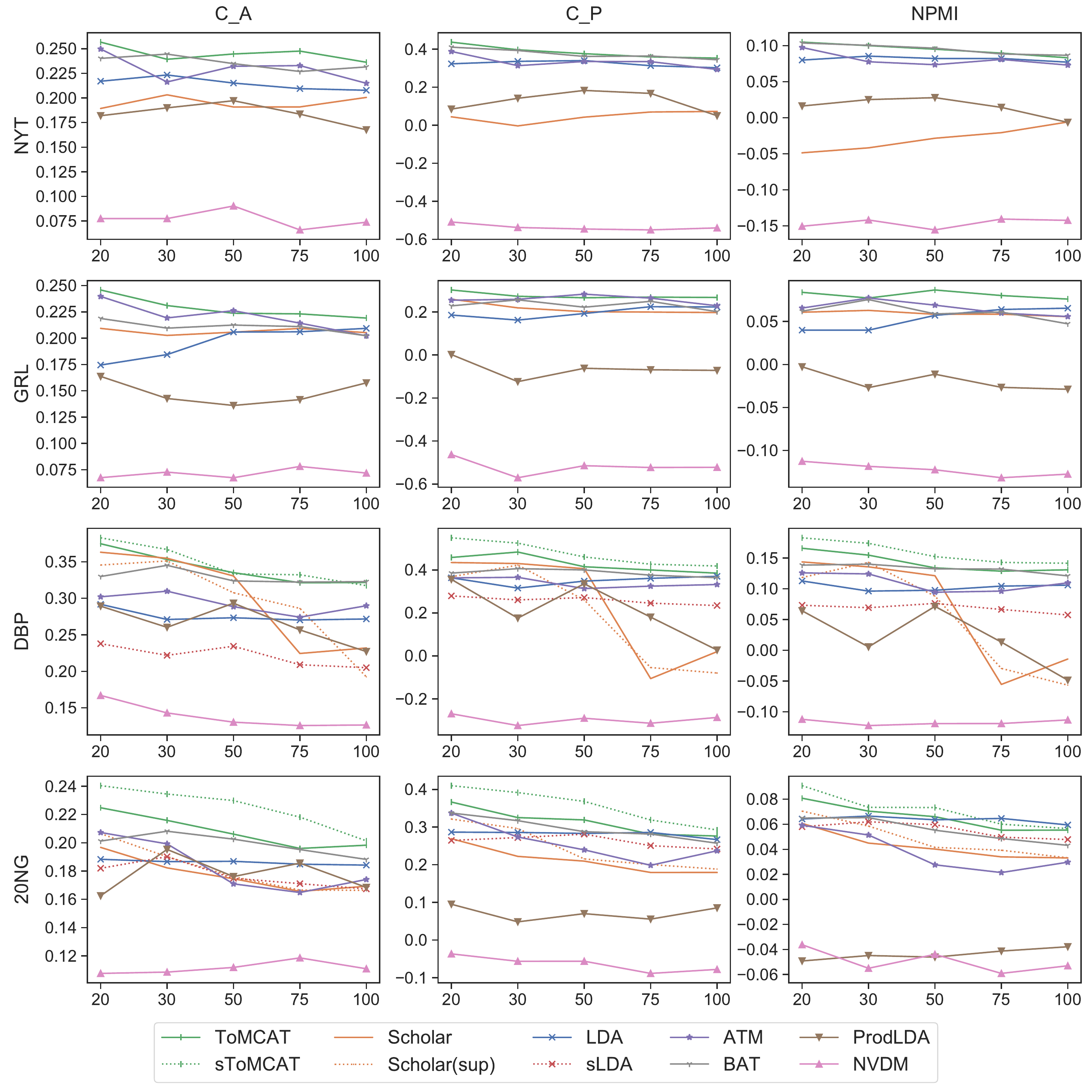}
  \caption{
    Topic coherence (C\_A, C\_P, NPMI) w.r.t. topic numbers on 4 datasets.
    Dotted lines denote supervised topic models.}
  \label{fig:line}
\end{figure*}

\setcounter{table}{2}
\begin{table}[h]
  \centering
  \resizebox{\columnwidth}{!}{
    \begin{tabular}{lllll}
      \toprule
      \bt{Vehicle} & \bt{Election} & \bt{Court}   & \bt{Fashion} \\
      \midrule
      car          & voter         & court        & fashion      \\
      tires        & poll          & lawsuit      & designer     \\
      fuel         & campaign      & case         & leather      \\
      driver       & percent       & ruling       & wear         \\
      truck        & primary       & antitrust    & dress        \\
      vehicle      & republican    & suit         & clothes      \\
      vehicles     & vote          & plaintiff    & skirt        \\
      gas          & democratic    & judge        & white        \\
      gasoline     & states        & settlement   & shirt        \\
      engine       & democrat      & federal      & pant         \\
      \midrule
      \bt{Cooking} & \bt{Baseball} & \bt{Disease} & \bt{Art}     \\
      \midrule
      cup          & inning        & patient      & artist       \\
      tablespoon   & run           & cancer       & painting     \\
      pepper       & hit           & doctor       & art          \\
      teaspoon     & homer         & hospital     & collection   \\
      garlic       & game          & drug         & exhibition   \\
      sauce        & yankees       & disease      & photograph   \\
      onion        & pitcher       & medical      & museum       \\
      chopped      & season        & therapy      & images       \\
      add          & hitter        & surgery      & gallery      \\
      butter       & pitch         & treatment    & exhibit      \\
      \bottomrule
    \end{tabular}
  }
  \caption{8 topics discovered by our model on NYT.}
  \label{tb:topic8}
\end{table}
\setcounter{table}{4}

\subsubsection{Supervised Topic Modeling}
Supervised topic modeling aims to leverage available document labels to benefit topic modeling.
Therefore we only conduct experiments on labeled datasets, i.e., DBpedia and 20 Newsgroups.
The experimental results are shown on the right part of Table \ref{tb:coherence}.
We expect the topic extraction results would be improved with the incorporation of topic labels.
However, this is not always the case as shown in Table \ref{tb:coherence}.
The supervised Scholar outperforms its unsupervised version on 20 Newsgroups
but the unsupervised one achieves higher coherence scores on DBpedia.
While sLDA fails to surpass its unsupervised counterpart LDA on both DBpedia and 20 Newsgroups.
On the contrary, improvements of sToMCAT over the unsupervised ToMCAT can be observed under all settings.
The results show that the incorporation of the supervised information seems to be
more effective in our proposed model,
probably contributing to the gradient-based loss balancing mechanism.
Overall, our model consistently outperforms sLDA and Scholar on all datasets and all topic coherence measures.

\subsubsection{Impact of Topic Numbers}
To investigate how topic coherence scores vary with respect to different topic number settings,
we show in Figure \ref{fig:line} the topic coherence measures on four datasets for all models.
Although there are exceptions that some baselines achieve higher scores
on specific experimental settings, the general conclusion is that
our models perform the best in both unsupervised and supervised topic modeling tasks.
On DBpedia and 20 Newsgroups datasets, sToMCAT consistently outperforms ToMCAT,
indicating the additional supervision helps generate more coherent topics.
We also notice that although the topic coherence measures of our models remain relatively stable
across topic numbers, there are slight drops on the DBpedia and 20 Newsgroups datasets
when the topic number becomes bigger.
This phenomenon may result from the fact that DBpedia and 20 Newsgroups datasets are less diverse than others.
There are only 14 and 20 categories in DBpedia and 20 Newsgroups datasets, respectively.
When the topic number is much larger than the ground-truth category number,
discriminating different topics would be more challenging.
Nevertheless, the overall superiorities of our models are significant as in Figure \ref{fig:line}.

\subsection{Text Classification}\label{ssec:expClass}
We now report text classification results of supervised topic models
: sLDA, Scholar, and sToMCAT.
To show that our model can learn both coherent and informative topics concurrently,
we use the same models as in the topic modeling experiments to classify test set documents,
and do not perform any further fine-tuning.
In our experiments, we found that the text classification performance
is influenced by topic numbers.
Therefore we conduct experiments with five topic number settings: 20, 30, 50, 75, and 100.

\setcounter{table}{3}
\begin{table*}[!t]
  \centering
  \resizebox{2\columnwidth}{!}{
    \begin{tabular}{llRRRRRRRRR}
      \toprule
      Dataset & Model
              &
      \multicolumn{1}{c}{20 ($\uparrow$)}
              &
      \multicolumn{1}{c}{30 ($\uparrow$)}
              &
      \multicolumn{1}{c}{50 ($\uparrow$)}
              &
      \multicolumn{1}{c}{75 ($\uparrow$)}
              &
      \multicolumn{1}{c}{100 ($\uparrow$)}
              &
      \multicolumn{1}{c}{Min ($\uparrow$)}
              &
      \multicolumn{1}{c}{Avg ($\uparrow$)}
              &
      \multicolumn{1}{c}{Max ($\uparrow$)}
              &
      \multicolumn{1}{c}{$\Delta$ ($\downarrow$)}                                                                                            \\
      \midrule
      \multirow{3}{*}{DBP}
              & sLDA    & 0.871      & 0.906      & 0.909      & 0.918      & 0.922      & 0.871      & 0.905      & 0.922      & 0.051      \\
              & Scholar & 0.949      & \bm{0.951} & 0.948      & 0.920      & 0.900      & 0.900      & 0.934      & 0.951      & 0.051      \\
              & sToMCAT & \bm{0.951} & \bm{0.951} & \bm{0.953} & \bm{0.936} & \bm{0.928} & \bm{0.928} & \bm{0.944} & \bm{0.953} & \bm{0.025} \\
      \midrule
      \multirow{3}{*}{20NG}
              & sLDA    & 0.529      & 0.572      & 0.563      & 0.608      & 0.613      & 0.529      & 0.577      & 0.613      & 0.084      \\
              & Scholar & 0.523      & 0.576      & 0.598      & \bm{0.617} & 0.610      & 0.523      & 0.585      & 0.617      & 0.094      \\
              & sToMCAT & \bm{0.642} & \bm{0.628} & \bm{0.616} & \bm{0.617} & \bm{0.616} & \bm{0.616} & \bm{0.624} & \bm{0.642} & \bm{0.026} \\
      \bottomrule
    \end{tabular}
  }
  \caption{Classification accuracy of supervised topic models
    with different topic numbers (20, 30, 50, 75, 100). `Min/Avg/Max' shows the minimum/average/maximum accuracy
    among different topic numbers.
    `$\Delta$' shows the variance of the classification accuracy across different topic numbers.
  }
  \label{tb:acc}
\end{table*}

Classification results are presented in Table \ref{tb:acc}.
We can see that our model not only achieves the best overall performance (the Max and Avg column),
but also has the highest accuracies on all dataset and topic number settings.
Compared to Scholar, our model achieves a slightly higher accuracy on DBpedia
and an accuracy improvement of $2.5\%$ on 20 Newsgroups.
The performance gain of our model over sLDA is more significant.
In addition to better classification results,
our model is also more robust to the change of topic numbers (the $\Delta$ column).
With the topic number increasing from $20$ to $100$,
the variance of the classification accuracy of our model is only $0.025$ and $0.026$
on DBpedia and 20 Newsgroups respectively, which is much lower than that of sLDA and Scholar.

\section{Conclusion}\label{sec:conclusion}
We have presented ToMCAT, a neural topic model with adversarial and cycle-consistent objectives,
and its supervised extension, sToMCAT.
ToMCAT employs a generator to capture semantic patterns in topics
and an encoder to encode documents into their corresponding topics.
sToMCAT further incorporates document labels into topic modeling.
The effectiveness of ToMCAT and sToMCAT is verified by experiments on topic modeling and text classification.
In the future, we plan to extend our model to cope with external word or document semantics.
It would also be interesting to explore alternative architectures other than CycleGAN
under our formulation of topic modeling.

\section*{Acknowledgments}
The authors would like to thank the anonymous
reviewers for insightful comments and helpful suggestions.
This work was funded in part by the National Key Research and Development Program
of China (2016YFC1306704) and the National Natural Science Foundation of China (61772132).

\bibliographystyle{acl_natbib}
\bibliography{tomcat}

\appendix

\section{Discovered Topics on NYTimes}
To gain an insight into the extracted topics,
we present the full list of 50 topics on NYTimes discovered by ToMCAT in Table \ref{appendix:tbl:tomcatTopic}.
As a comparison, topics discovered by LDA
are shown in Table \ref{appendix:tbl:LDATopic}.

\begin{table*}[!b]
  \centering
    \begin{tabular}{llRRRRRRRR}
      \toprule
      stock fund firm companies online investment investor broker company customer                           \\
      car tires fuel driver truck vehicle vehicles gas gasoline engine                                       \\
      voter poll campaign percent primary republican vote democratic states democrat                         \\
      building project apartment house houses homes resident housing estate square                           \\
      flight passenger airline plane airport customer carrier pilot airlines tires                           \\
      yard game touchdown play team quarterback season goal offense pass                                     \\
      show actor producer character series film network award television comedy                              \\
      computer www user site web window file com files mail                                                  \\
      court lawsuit case ruling antitrust suit plaintiff judge settlement federal                            \\
      officer police investigation mayor prosecutor department charges complaint criminal official           \\
      film movie character actor movies director comedy script minutes scenes                                \\
      company merger companies billion deal cable stock acquisition network market                           \\
      union worker employees company job contract pay employer manager benefit                               \\
      school student teacher test program district children education percent parent                         \\
      friend course article black guy thought movie husband film house                                       \\
      fashion designer leather wear dress clothes skirt white shirt pant                                     \\
      black white protester flag town crowd street protest community school                                  \\
      music album song band jazz artist rock guitar musical singer                                           \\
      campaign political money fund president governor presidential republican election candidates           \\
      boy father cuban family relatives mother child son custody grandmother                                 \\
      computer privacy software companies information web user sites internet company                        \\
      jet coach patriot season player team draft coaching defensive football                                 \\
      cell genome scientist genes human genetic researcher gene disease study                                \\
      died survived degree film graduated served wife student born article                                   \\
      rebel military war soldier troop attack terrorist civilian forces bombing                              \\
      abortion religious conservative support european conservatives government republican thunderstorm vote \\
      cup tablespoon pepper teaspoon garlic sauce onion chopped add butter                                   \\
      com commentary daily tduncan information toder holiday eta staffed sport                               \\
      gun gay women firearm law violence sexual percent bill shooting                                        \\
      war church government country african nation communist black priest leader                             \\
      drug missile nuclear weapon defense official sanction administration missiles countries                \\
      forest bird fire species water land fish fires animal acres                                            \\
      book memoir author bookstores fiction ages nonfiction writer reader witchcraft                         \\
      palestinian israeli peace israelis jewish syrian violence arab summit lebanese                         \\
      ballot recount votes election vote counties county count board manual                                  \\
      race medal racing meter gold team track driver races lap                                               \\
      tournament fight round par match tour game champion fighter golf                                       \\
      percent survey population economy immigrant economic million companies worker wage                     \\
      inning run hit homer game yankees pitcher season hitter pitch                                          \\
      fax syndicate www tour com hotel trip ticket room telex                                                \\
      penalty death execution prosecutor murder trial jury inmates prison lawyer                             \\
      convention speech party campaign republican democratic delegates president presidential democrat       \\
      patient cancer doctor hospital drug disease medical therapy surgery treatment                          \\
      point game team shot rebound pointer foul guard minutes play                                           \\
      tax taxes bill cut surplus income plan proposal spending billion                                       \\
      stock percent quarter earning market index company analyst cent investor                               \\
      election party government opposition political minister power president country leader                 \\
      campaign debate debates candidates presidential aides president vice reporter adviser                  \\
      artist painting art collection exhibition photograph museum images gallery exhibit                     \\
      yankees team fan player baseball game games football league stadium                                    \\
      \bottomrule
    \end{tabular}
  \caption{Full list of 50 topics on NYTimes discovered by ToMCAT.}
  \label{appendix:tbl:tomcatTopic}
\end{table*}

\begin{table*}[tb]
  \centering
    \begin{tabular}{llRRRRRRRR}
      \toprule
      official agency investigation letter statement comment public office document interview               \\
      \emph{percent women number according study survey group found likely million}                         \\
      political government president power leader country party election opposition minister                \\
      building project home local town resident area center million house                                   \\
      scientist human research cell science researcher found called brain light                             \\
      country foreign trade countries government nation european economic american international            \\
      \emph{need problem feel look right hard happen help trying change}                                    \\
      film movie character play actor movies director minutes cast role                                     \\
      election vote ballot votes voter count recount result hand campaign                                   \\
      car driver seat truck drive driving road vehicle model wheel                                          \\
      company companies million business firm deal industry billion executive market                        \\
      job worker employees union manager president working contract member pay                              \\
      customer sales sell product buy consumer business price market store                                  \\
      water bird fish weather rain animal wind plant land trees                                             \\
      school student program teacher college high education class children public                           \\
      \emph{guy tell look kid bad big dog right real word}                                                  \\
      court case law decision lawyer federal legal judge right lawsuit                                      \\
      point game play team goal shot games lead left half                                                   \\
      look show art collection fashion designer artist style wear painting                                  \\
      round won sport fight shot player final tournament gold event                                         \\
      drug patient doctor medical health cancer hospital disease treatment care                             \\
      book author writer writing wrote read published magazine find write                                   \\
      campaign republican president presidential democratic voter political candidates candidate convention \\
      meeting official talk agreement deal leader decision conference president negotiation                 \\
      \emph{article special fax information syndicate contact visit buy separate purchased}                 \\
      computer system software technology user program digital window internet access                       \\
      com question daily newspaper american today information business sport statesman                      \\
      police death officer case crime prison criminal prosecutor trial victim                               \\
      military system security defense nuclear weapon official administration attack arm                    \\
      web site com www sites mail online information internet telegram                                      \\
      \emph{word fact sense question perhap course point matter mean view}                                  \\
      black group white religious right gay church jewish member flag                                       \\
      money million tax plan pay billion cut cost fund program                                              \\
      flight plane ship crew pilot air passenger boat airport hour                                          \\
      war palestinian peace soldier israeli military troop violence attack killed                           \\
      history century french known german today american ago died modern                                    \\
      room house wall door floor hand water window light inside                                             \\
      oil prices plant million gas production energy industry power cost                                    \\
      family father children son mother boy home child parent daughter                                      \\
      percent stock market fund quarter growth economy investor earning analyst                             \\
      room hotel trip restaurant tour travel night visit visitor dinner                                     \\
      \emph{night crowd hour morning reporter hand street told moment left}                                 \\
      race won win run track winner running racing place winning                                            \\
      music song band sound record album musical show pop rock                                              \\
      cup food minutes add oil tablespoon fat chicken large pepper                                          \\
      bill group law gun support legislation issue member right federal                                     \\
      run hit game season inning yankees home baseball right games                                          \\
      \emph{wanted thought told friend asked knew took felt saw ago}                                        \\
      show television network media station commercial series radio viewer rating                           \\
      team player season game play coach yard games football league                                         \\
      \bottomrule
    \end{tabular}
  \caption{Full list of 50 topics on NYTimes discovered by LDA.}
  \label{appendix:tbl:LDATopic}
\end{table*}

\end{document}